\documentclass[a4paper,fleqn]{cas-sc}

\usepackage[numbers]{natbib}

\usepackage{multirow} 
\usepackage{makecell} 
\usepackage{rotating}
\usepackage[percent]{overpic}
\usepackage{float}
\usepackage{graphicx}
\usepackage{subcaption}
\usepackage{tikz}
\usepackage{placeins}   

\def\tsc#1{\csdef{#1}{\textsc{\lowercase{#1}}\xspace}}
\tsc{WGM}
\tsc{QE}

\usepackage{placeins}

\begin{document}


\shorttitle{}    

\shortauthors{}

\title [mode = title]{ByteAction: Byte-space Action Recognition Foundation Model}

\author[1]{Fangcheng~Li}
\credit{Writing – original draft, Methodology, Formal analysis, Data curation, Visualization}
\author[1]{Zhen~Yu}
\credit{Writing – original draft, Validation, Investigation}
\author[1]{Kejun Wu}
\credit{Writing – review \& editing, Methodology, Funding acquisition}
\cormark[1]
\ead{kjwu@hust.edu.cn}

\author[1]{Qiong Liu}
\credit{Supervision, Project administration}         

\author[1]{You Yang}
\credit{Supervision, Resources}

\affiliation[1]{organization={School of Electronic Information and Communications, Huazhong University of Science and Technology},
                city={Wuhan},
                postcode={430074}, 
                country={China}}

\begin{abstract}
Byte-space Action Recognition (BAR) aims to recognize human actions directly from compressed image bitstreams without any pixel decoding. By operating entirely in byte space, BAR is inherently independent of file integrity and pixel-level reconstruction, making it naturally applicable to privacy-sensitive scenarios and robust against bitstream corruption. In this paper, we propose ByteAction, a BAR foundation model that achieves accurate action recognition on corrupted image bitstreams. ByteAction follows a dual-view byte-level recognition framework. It constructs weakly and strongly corrupted bitstream views, which are augmented by Bitstream Pattern Augmentation (BPA) and encoded with a shared ByteFormer backbone. The model is optimized with both classification and corruption consistency objectives. Specifically, we propose Bitstream Pattern Augmentation (BPA), which reshapes one-dimensional byte sequences into two-dimensional byte matrix and applies region-level erasure to encourage the model to learn robust cross-region byte dependencies. We further propose a Corruption Consistency Training strategy that constrains the model to maintain stable predictions across different corruption severities through bidirectional KL divergence. Experiments on the image bitstream from Stanford40, PPMI, and PASCAL VOC 2012 Action demonstrate that ByteAction achieves state-of-the-art corruption robustness across all scenarios while maintaining competitive intact bitstream performance.

\end{abstract}


\begin{highlights}
\item We propose ByteAction for Byte-space Action Recognition
\item BPA diversifies bitstream patterns through 2D byte-matrix erasure.
\item Consistency training stabilizes predictions across corruption severities.
\item ByteAction outperforms baselines on three corrupted bitstream datasets.
\end{highlights}

\begin{keywords}
Foundation models \sep Action Recognition \sep corrupted bitstream \sep deep learning 
\end{keywords}

\maketitle

\section{Introduction}\label{sec:intro}
Action recognition is a fundamental task in computer vision, with broad applications in intelligent surveillance, human-computer interaction, and healthcare monitoring \cite{Alturki2020HAR, Li2019Gesture}. Recent convolutional neural networks and Vision Transformers have achieved strong performance by learning discriminative representations from decoded RGB images \cite{Moutsis2023HAR}. These pixel-domain recognition pipelines implicitly assume that compressed image files can be successfully decoded into valid visual inputs. However, in real-world storage and transmission, compressed image bitstreams are often exposed to corruption. Such corruption may degrade the decoded image quality or even cause complete decoding failure, making pixel-domain action recognition pipelines unreliable. 

To overcome the obstacles posed by bitstream corruption, existing studies have mainly explored two research directions: pixel-domain restoration and bitstream-domain understanding. Pixel-domain restoration methods, including advanced image super-resolution \cite{liu2026promptsr} and image inpainting \cite{yeh2024image}, attempt to recover degraded or missing visual content before recognition. However, they still require the corrupted file to be at least decodable \cite{Liu2023JPEGRestore}. Once the bitstream cannot be successfully decoded, these methods fail to provide valid visual inputs for downstream action recognition.
Bitstream-domain methods avoid direct RGB reconstruction by exploiting information from compressed files. Early bitstream-domain methods performed recognition based on compression-domain representations, such as DCT coefficients.These methods avoid full RGB reconstruction, but they still require valid bitstream parsing to extract compression-domain features. When bitstream corruption prevents such features from being reliably extracted, they become ineffective for downstream recognition \cite{Dong2023Survey}. To fully remove the dependence on decoders, recent studies have started to model raw byte streams directly. Representative byte-level models, such as ByteFormer \cite{horton2023bytes} and bGPT \cite{Wu2024bGPT}, demonstrate the feasibility of visual understanding in byte space.
Nevertheless, these models are mostly designed for intact bitstreams, resulting in degraded performance under corrupted inputs.

\begin{figure}
    \centering
    \begin{subfigure}[b]{0.45\textwidth}
        \centering
        \includegraphics[width=\textwidth]{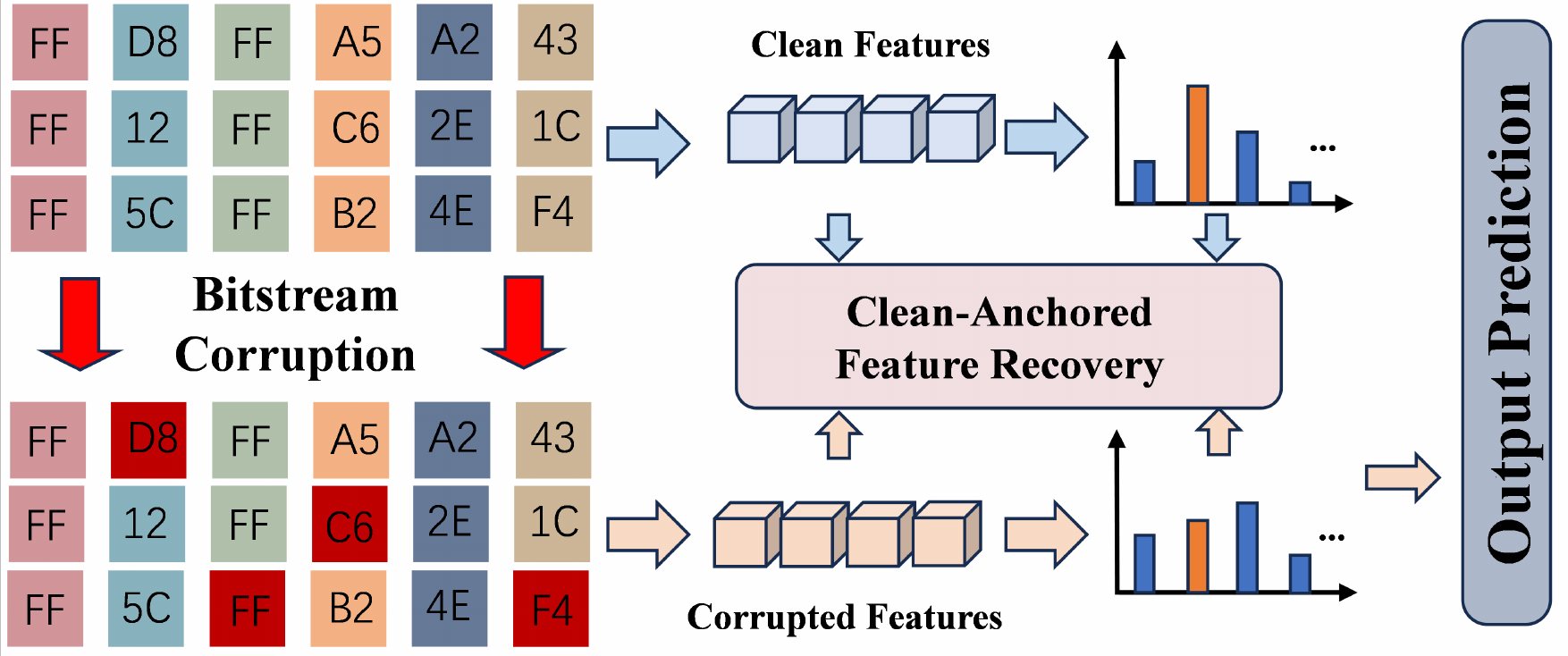} 
        \label{fig:pipeline_a}
    \end{subfigure}%
    \hfill
    \begin{tikzpicture}
        \draw[dashed, line width=0.8pt, gray] (0,0) -- (0,4.5); 
    \end{tikzpicture}%
    \hfill
    \begin{subfigure}[b]{0.46\textwidth}
        \centering
        \includegraphics[width=\textwidth]{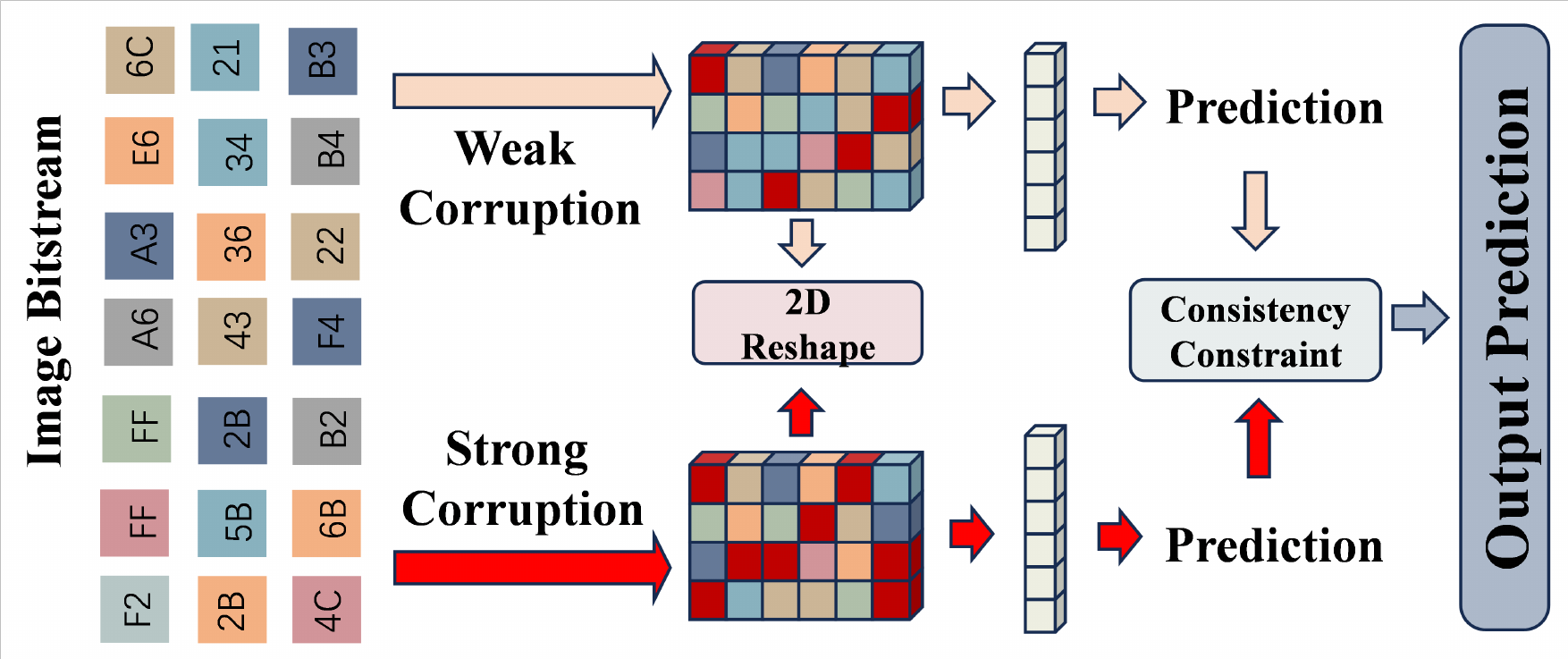} 
        \label{fig:pipeline_b}
    \end{subfigure}
    \caption{Comparison of two bitstream-domain pipelines for action recognition under bitstream corruption. (a) The existing approach aligns corrupted features with intact features through a frozen teacher model. (b) The proposed ByteAction reshapes byte sequences into two-dimensional matrices via BPA and enforces prediction consistency between two corruption levels without requiring a teacher model.}
    \label{fig:pipeline_comparison}
\end{figure}
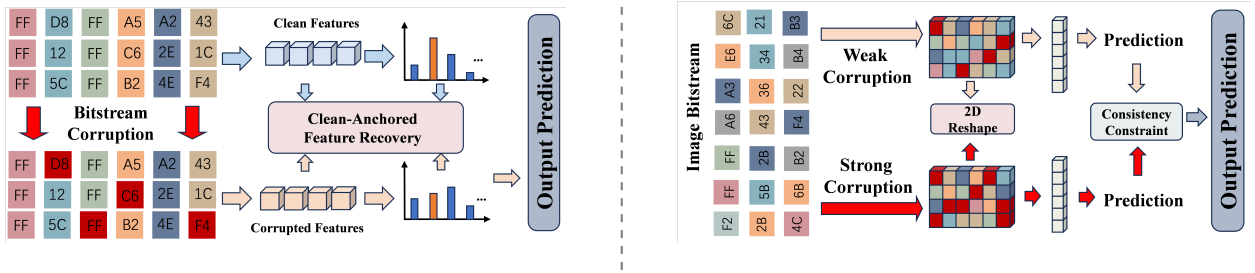

These limitations have motivated the recently introduced Byte-space Action Recognition (BAR) task~\cite{li2026bitstream}, which aims to recognize human actions directly from compressed image bitstreams without relying on pixel decoding. However, BAR remains highly challenging. Image bitstreams are long and discrete one-dimensional byte sequences, in which action-related semantics are implicitly encoded rather than explicitly represented as visual patterns. This makes it difficult for models to capture discriminative action representations directly from raw bytes. Meanwhile, image bitstreams may suffer from unpredictable corruption during storage and transmission. Such corruption can damage useful byte patterns, making it difficult for the model to produce stable predictions.
Moreover, as illustrated in Fig.~\ref{fig:pipeline_comparison}(a), the existing BRACE framework learns corruption robustness by aligning corrupted representations with intact anchors through an additional frozen branch. Although effective, this intact-anchored paradigm requires explicit clean-corrupted pairing during training. These challenges highlight the need for a simpler teacher-free BAR framework that can directly learn robust representations from corrupted bitstreams.

Motivated by these challenges, we propose ByteAction, a foundation model for Byte-space Action Recognition (BAR), which directly recognizes human actions from corrupted image bitstreams without relying on pixel decoding. In contrast to the intact-anchored paradigm, ByteAction learns directly from two corrupted views with different severity levels, as illustrated in Fig.~\ref{fig:pipeline_comparison}(b). Specifically, we introduce Bitstream Pattern Augmentation (BPA), which reshapes one-dimensional byte sequences into two-dimensional byte matrices and applies region-level erasure to diversify bitstream patterns. We further introduce Corruption Consistency Training, which encourages stable predictions across different corruption severities through bidirectional KL divergence.
The main contributions of this paper are as follows:
\begin{itemize}
    \item We reveal that image bitstreams contain exploitable byte-layout patterns after two-dimensional reshaping. Based on this observation, we propose Bitstream Pattern Augmentation (BPA), which applies region-level erasure in the reshaped byte matrix to diversify bitstream patterns and improve robust byte-level representation learning.
    \item We propose a Corruption Consistency Training strategy that constrains the model to produce stable predictions across different corruption severities via bidirectional KL divergence, requiring no additional models or intact references during training.
    \item Extensive experiments on the bitstream from Stanford40, PPMI, and PASCAL VOC 2012 Action demonstrate that ByteAction achieves state-of-the-art corruption robustness across diverse corruption scenarios while maintaining competitive performance on intact bitstream .
\end{itemize}

\section{Related Work}\label{sec:intro}

\subsection{Action Recognition}

Still-image action recognition focuses on inferring human actions from spatial cues without temporal information. Early paradigms predominantly relied on hand-crafted features, human pose estimation, and part-based detectors. For instance, Bourdev et al.~\cite{Bourdev2009Poselets} utilized Poselets to encode local body configurations, while subsequent works explicitly modeled the spatial constraints between human poses and object affordances~\cite{Gupta2009HOI,Yao2010HOI}. The advent of deep learning shifted this paradigm toward end-to-end feature extraction. Region-based approaches, such as the framework proposed by Gkioxari et al.~\cite{Gkioxari2015RCNN}, effectively localized action-specific regions. More recently, attention-based architectures such as ConViT~\cite{Hosseyni2024ConViT} have been employed to adaptively capture action-relevant spatial contexts, significantly improving recognition accuracy.

To achieve a more comprehensive semantic understanding of visual scenes, human-object interaction (HOI) detection extends traditional action recognition by jointly predicting the human, the interacting object, and their relationship. Driven by large-scale benchmarks such as HICO~\cite{Chao2015HICO}, multi-branch neural architectures, such as InteractNet~\cite{Gkioxari2018HOI}, emerged as a standard paradigm by decoupling human, object, and interaction feature extraction. More recently, vision-language foundation models have been increasingly adopted for open-vocabulary interaction understanding. For example, Cao et al.~\cite{Cao2023HOI} leveraged foundation models to address long-tail interaction distributions. Complementing this direction, Gao et al.~\cite{gao2024contextual} explored contextual human-object interaction understanding by leveraging knowledge from a pre-trained large language model. Building on these advances in cross-modal semantic modeling, recent approaches such as soft-label guided multi-granularity prompting~\cite{han2026soft} have further refined the extraction of fine-grained interaction representations.

A major limitation of these methods is their strict reliance on fully decoded visual inputs. When bitstream corruption prevents reliable image reconstruction, these pixel-based methods cannot extract valid visual features, rendering them ineffective for downstream action recognition.

\subsection{Bitstream Understanding}
In visual understanding tasks, the standard pipeline requires fully decoding compressed files into pixel representations before performing any analysis, which introduces significant computational overhead and exposes the raw image content. To address these limitations, researchers have explored performing analysis directly on compressed data or raw bitstreams at different levels of abstraction. 
In the image domain, early works directly utilized compressed data to avoid full decoding. For instance, Jamil et al.~\cite{Jamil2019CBIR} extracted DCT coefficients for image retrieval, Dugad et al.~\cite{Dugad2001Resize} and Mukherjee et al.~\cite{Mukherjee2002Resize} achieved efficient image resizing in the DCT domain, and Shen et al.~\cite{Shen1997Feature, Shen1996Edge} exploited the 8$\times$8 block structure to extract local features.While recent advancements have significantly improved deep image and video compression architectures \cite{wu2025end}, standard visual understanding methods still depend on correctly parsing the bitstream header, quantization tables, and entropy coding structures.

While these methods reduce computational cost by avoiding complete pixel reconstruction, they still depend on correctly parsing the bitstream header, quantization tables, and entropy coding structures. A related research direction is multimedia file fragment classification (MFFC), which aims to identify the types of file fragments directly from raw bytes without relying on file headers or metadata. Early methods such as Sceadan \cite{BeebeMLS13} relied on hand-crafted statistical features and traditional classifiers, while more recent approaches such as FiFTy \cite{MittalKM21} adopted 1D CNNs to learn representations directly from byte sequences, eliminating the need for manual feature engineering. Beyond modeling each sector independently, Wang et al. \cite{wang2024intra} introduced a joint self-attention network to capture intra-sector byte dependencies and inter-sector contextual information across neighboring sectors. ByteNet \cite{liu2024bytenet} further introduced a visual perspective by reshaping one-dimensional byte streams into two-dimensional representations and applying Vision Transformers to capture spatial byte correlations. Collectively, these studies demonstrate that raw byte sequences contain meaningful structural patterns that can be effectively exploited by learning-based methods.

More recently, byte-level models have extended this idea to 
end-to-end visual understanding. bGPT \cite{Wu2024bGPT} 
treats diverse data modalities uniformly as byte sequences 
and models them with Transformers through next-byte 
prediction, demonstrating the generality of byte-level 
representations. ByteFormer \cite{horton2023bytes} 
applies this paradigm specifically to visual recognition by 
taking binary streams of compressed image files as input via 
learnable byte embeddings, achieving competitive performance 
on standard image classification benchmarks. To address the 
high computational cost of processing long byte sequences, 
hierarchical chunking \cite{pagnoni2025byte} and dynamic 
token merging \cite{Kallini2025MrT5} have been proposed to 
reduce effective sequence length while preserving key 
information.However, existing byte-level models are primarily designed for intact bitstream inputs, and their performance often degrades substantially under bitstream corruption.

\subsection{Multimodal Foundation Models}
With the rapid advancement of deep learning, AI systems are 
increasingly required to process and integrate data from 
diverse modalities including text, images, video, audio, and 
sensor signals. Multimodal foundation models pretrained on 
large-scale cross-modal data have demonstrated strong 
transferability and generalization, driving progress in tasks 
such as visual question answering, image captioning, and 
cross-modal retrieval \cite{lu2025representation}.

Among these, vision-language models have received 
particular attention due to the complementary nature of 
visual and textual information. CLIP~\cite{radford2021clip} 
learns aligned representations of images and text through 
contrastive pretraining on large-scale paired data, and its 
text encoder has been widely adopted for zero-shot 
recognition and semantic knowledge transfer in downstream 
tasks. Building on this foundation, multimodal large language 
models such as LLaVA~\cite{liu2023visual} and 
InternVL~\cite{chen2024internvl} integrate visual perception 
with the reasoning capabilities of large language models, 
achieving notable progress in open-domain visual 
understanding. Researchers have also explored extending 
foundation model capabilities to non-traditional modalities and various tasks.LLMArk~\cite{wu2026llmark} adapts foundation models for instance-aware flood risk assessment.MCF-LLM~\cite{xu2026mcf} incorporates textual prompts 
with spatio-temporal traffic data through cross-modal fusion 
for traffic flow prediction.Furthermore, large language models have been leveraged for fault diagnosis in rotating machinery by processing discretized signal representations~\cite{zhang2025fault}.The remarkable success of these foundation models across diverse fields inspires the possibility of treating compressed image bitstreams as a unique and specialized modality, motivating the exploration of byte-space visual understanding.

\section{The Proposed ByteAction}\label{sec:method}

\subsection{Overview}\label{subsec:overview}

 \begin{figure}
    \centering
    \includegraphics[width=1\linewidth]{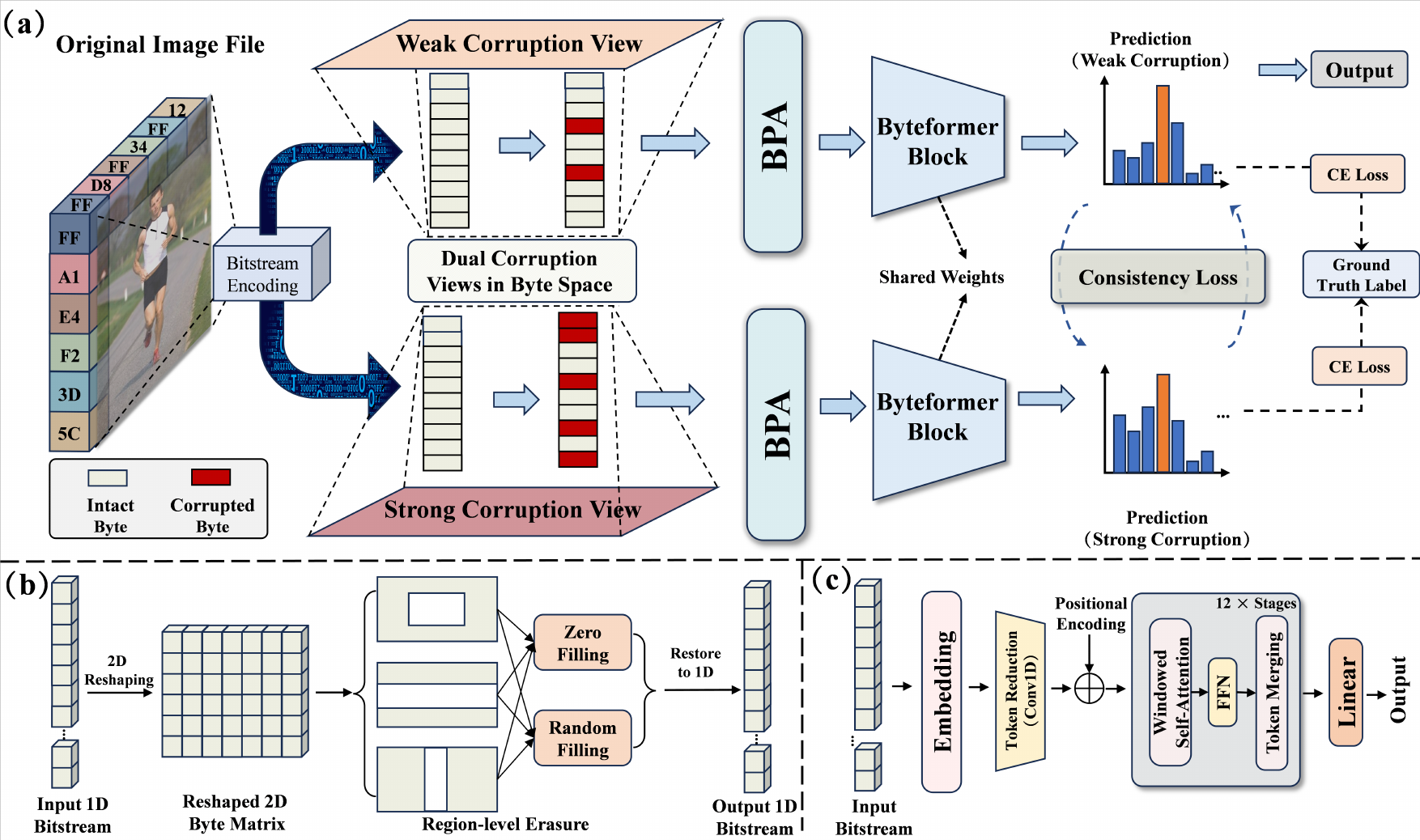} 
    \caption{Overview of the proposed ByteAction framework. (a) ByteAction constructs dual corruption views in byte space from the input image bitstream, including weakly and strongly corrupted views. The two views are augmented by Bitstream Pattern Augmentation (BPA), encoded by a shared-weight ByteFormer backbone, and optimized with both cross-entropy classification losses and a prediction consistency loss. At inference time, only a single bitstream input and one forward pass are required.(b) Detail of the Bitstream Pattern Augmentation (BPA) module: the 1D byte sequence is reshaped into a 2D matrix, region-level erasure is applied with either zero filling or random filling, and the result is restored to a 1D sequence.(c) Simplified ByteFormer structure, consisting of byte embedding, token reduction, positional encoding, stacked self-attention and feed-forward stages with token merging, followed by a linear prediction head.}
    \label{fig:framework} 
\end{figure}

Fig.~\ref{fig:framework} illustrates the overall framework of ByteAction. Given a training image, we first encode it into an image bitstream. We then employ the Real-world Bitstream Corruption Simulator (RBCS)~\cite{li2026bitstream} as the corruption operator to generate two byte views with different corruption severities. Specifically, the corruption parameters are independently sampled from a weak range and a strong range, producing a weakly corrupted view and a strongly corrupted view, respectively. During training, each corrupted view is further processed by the proposed Bitstream Pattern Augmentation (BPA) module. BPA reshapes the one-dimensional byte sequence into a two-dimensional byte matrix. It then applies region-level erasure and flattens the result back into a one-dimensional sequence.
The two augmented byte sequences are fed into a shared-weight ByteFormer backbone. Each branch produces an action prediction. The training objective includes classification losses for both branches and a consistency loss between their predictions. This consistency loss encourages ByteAction to produce stable predictions under different corruption severities.At inference time, ByteAction only requires a single bitstream input and one forward pass. Therefore, the dual-view training design does not introduce additional inference cost.

\begin{figure}
    \centering
    \begin{subfigure}[b]{0.32\textwidth}
        \centering
        \includegraphics[width=\textwidth]{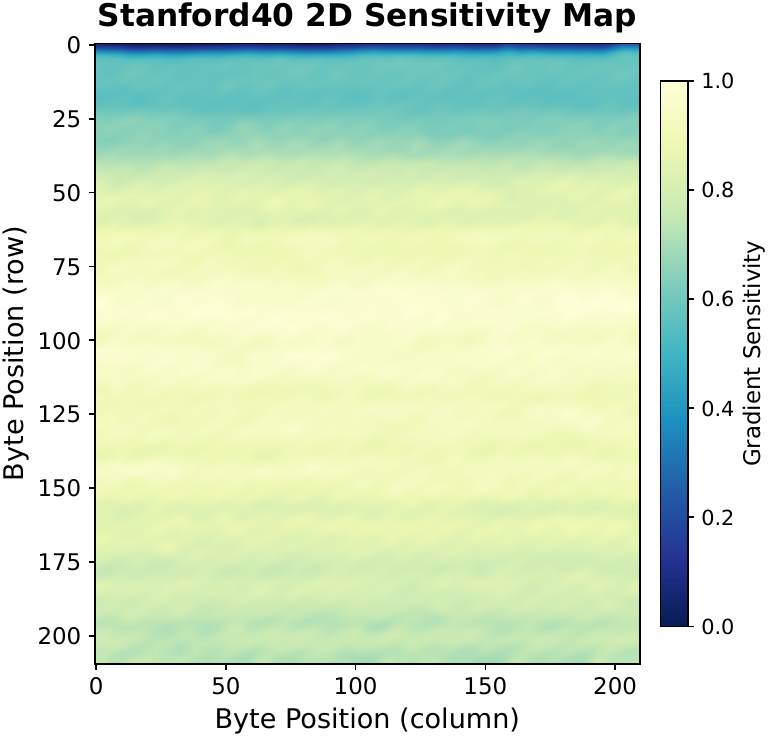}
        \caption{} 
        \label{fig:sub1}
    \end{subfigure}
    \hfill
    \begin{subfigure}[b]{0.32\textwidth}
        \centering
        \includegraphics[width=\textwidth]{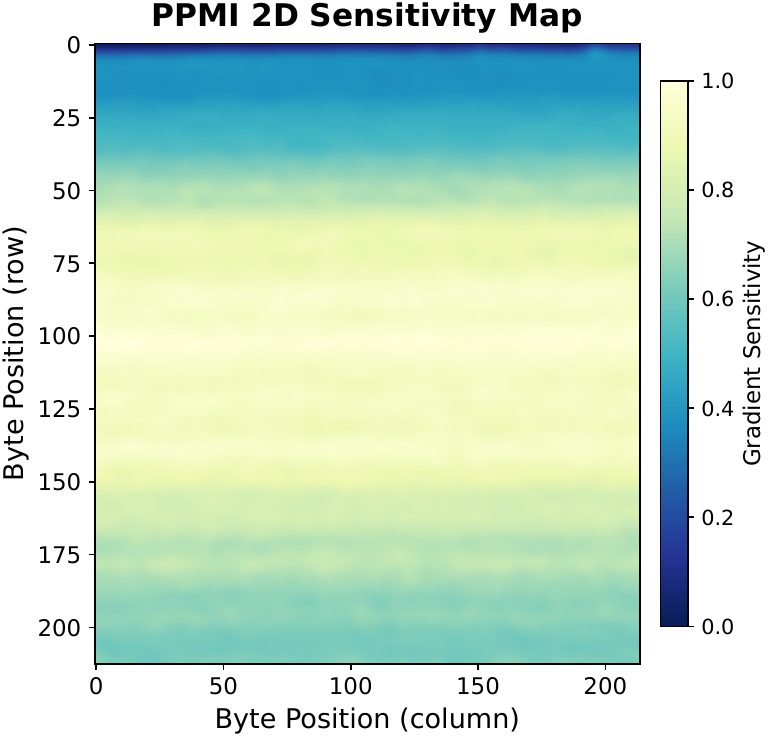}
        \caption{} 
        \label{fig:sub2}
    \end{subfigure}
    \hfill
    \begin{subfigure}[b]{0.32\textwidth}
        \centering
        \includegraphics[width=\textwidth]{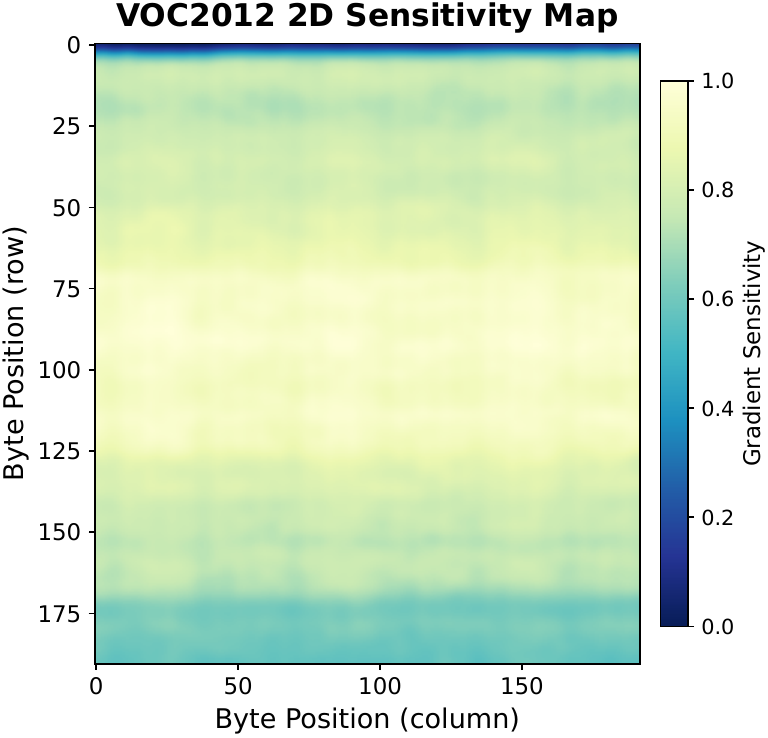}
        \caption{} 
        \label{fig:sub3}
    \end{subfigure}   
    
    \caption{Two-dimensional gradient sensitivity maps on (a) Stanford40, (b) PPMI, and (c) VOC2012 Action. The one-dimensional byte-level sensitivity sequence is reshaped into a two-dimensional matrix. Spatially coherent patterns indicate that the bitstream preserves an implicit two-dimensional structure inherited from the JPEG encoding order.}
    \label{fig:2d_sensitivity}
\end{figure}

\subsection{Bitstream Pattern Augmentation}\label{subsec:BPA}
Existing byte-level models usually treat JPEG bitstreams as flat one-dimensional sequences, ignoring possible structural dependencies among byte positions. However, JPEG encoding organizes image content in a block-wise scanning order, which suggests that the resulting byte sequence may still contain non-random local patterns. This motivates us to examine whether useful two-dimensional patterns can be exposed by reshaping the byte sequence.
To analyze this property, we conduct a gradient-based sensitivity analysis. For each test sample, we compute the gradient of the classification loss with respect to the output of the byte embedding layer, and use the $L_2$ norm at each byte position as its sensitivity score. The sensitivity scores are averaged over all test samples to obtain a dataset-level byte sensitivity sequence. We then reshape this one-dimensional sensitivity sequence into a two-dimensional byte matrix for visualization. As shown in Fig.~\ref{fig:2d_sensitivity}, the reshaped sensitivity maps on all three datasets consistently show higher sensitivity values around the central regions.
This observation suggests that JPEG bitstreams contain exploitable two-dimensional byte patterns after reshaping.
Based on the discovery, we propose Bitstream Pattern Augmentation (BPA). As illustrated in Fig.~\ref{fig:framework}(b), BPA consists of three steps: two-dimensional reshaping, region-level erasure , and one-dimensional restoration.
 
\noindent\textbf{Two-dimensional reshaping.}
Given a corrupted byte sequence of length $N'$, we compute the reshaping dimensions as $h = \lfloor \sqrt{N'} \rfloor$ and $w = \lfloor N' / h \rfloor$, and reshape the first $h \times w$ bytes into a two-dimensional matrix $\mathbf{M} \in \mathbb{R}^{h \times w}$. This operation approximately recovers the implicit spatial layout of the bitstream, so that a local rectangular region in the matrix roughly corresponds to a spatial area in the original image.

\noindent\textbf{Region-level erasure and filling.}
After reshaping the byte sequence into a two-dimensional matrix, BPA applies region-level erasure to generate diverse bitstream pattern variations. We consider three erasure modes: block erasure, horizontal stripe erasure, and vertical stripe erasure. Block erasure randomly removes a rectangular region whose height and width are sampled from $[r_{\min} \cdot h,\; r_{\max} \cdot h]$ and $[r_{\min} \cdot w,\; r_{\max} \cdot w]$, respectively. Horizontal and vertical stripe erasure remove consecutive rows or columns, with the stripe size sampled from $[s_{\min} \cdot h,\; s_{\max} \cdot h]$ or $[s_{\min} \cdot w,\; s_{\max} \cdot w]$. The erasure mode is randomly selected according to $p_{\text{stripe}}$, and each augmentation performs $K$ erasure operations to produce different regional missing patterns.For each erased region, BPA uses two filling strategies. With probability $p_r$, the region is filled with uniformly random byte values to simulate information corruption; otherwise, it is filled with zeros to simulate information loss. 

\noindent\textbf{One-dimensional restoration.}
After all erasure operations are completed, the two-dimensional matrix is flattened and concatenated with the remaining bytes to restore the one-dimensional sequence, which is then fed into the backbone network.

The key goal of BPA is to increase the diversity of bitstream patterns during training. Different from conventional byte-level augmentation that perturbs individual bytes independently, BPA reshapes the one-dimensional byte sequence into a two-dimensional byte matrix and then applies region-level erasure. This process generates diverse regional missing or corrupted patterns in the byte matrix, making the training samples more varied in byte space. As a result, the model is discouraged from relying on fragile local byte patterns and is encouraged to learn more stable byte-level representations from the remaining context.During training, BPA is applied to each corrupted byte sequence with probability $p_{\text{BPA}}$.

\subsection{Corruption Consistency Training}\label{subsec:consistency}

Prior approaches attempt to align corrupted representations with intact anchors. However, as corruption severity increases, the gap between corrupted and intact representations grows substantially, making accurate feature recovery increasingly difficult. We propose an alternative training objective based on a simpler principle: the same image should receive the same prediction regardless of how severely its bitstream is corrupted. Rather than demanding that the model reconstruct lost information, we only require that its predictions remain stable across different corruption levels.
The weak and strong corruption versions are generated online during training. For each sample, corruption parameters are independently sampled from two predefined ranges: the weak range produces mild degradation that preserves most of the byte structure, while the strong range introduces substantial byte-level distortion. This ensures the model is exposed to a wide spectrum of corruption intensities during training, and the consistency objective bridges the gap between the two extremes.

During training, each image is corrupted at two severity levels to produce a weak version $\widetilde{\mathbf{B}}_{\text{weak}}$ and a strong version $\widetilde{\mathbf{B}}_{\text{strong}}$. Both are processed by the same shared-weight network $\mathcal{F}_\theta$ to obtain prediction logits $\mathbf{z}_{\text{weak}}$ and $\mathbf{z}_{\text{strong}}$. The consistency loss is defined as the symmetrized KL divergence between the two softened prediction distributions:
\begin{equation}\label{eq:consistency}
\mathcal{L}_{\text{con}} = \frac{1}{2} \Big[ D_{\text{KL}}\big(\sigma(\mathbf{z}_{\text{weak}} / T) \;\|\; \sigma(\mathbf{z}_{\text{strong}} / T)\big) + D_{\text{KL}}\big(\sigma(\mathbf{z}_{\text{strong}} / T) \;\|\; \sigma(\mathbf{z}_{\text{weak}} / T)\big) \Big] \cdot T^2
\end{equation}
where $\sigma(\cdot)$ denotes the softmax function and $T$ is a temperature parameter that smooths the prediction distribution to expose richer inter-class similarity information. The symmetrized form ensures that both distributions are mutually constrained, avoiding the optimization bias that would arise from a single-direction divergence.

\subsection{Training Objective}\label{subsec:objective}

The total training loss combines classification losses on both branches with the consistency regularization:
\begin{equation}\label{eq:total}
\mathcal{L}_{\text{total}} = \mathcal{L}_{\text{CE}}(\hat{y}_{\text{weak}},\; y) + \mathcal{L}_{\text{CE}}(\hat{y}_{\text{strong}},\; y) + \lambda \cdot \mathcal{L}_{\text{con}}
\end{equation}
where $\mathcal{L}_{\text{CE}}$ denotes the cross-entropy loss with label smoothing, $\hat{y}_{\text{weak}}$ and $\hat{y}_{\text{strong}}$ represent the predicted class probabilities derived from the logits $\mathbf{z}_{\text{weak}}$ and $\mathbf{z}_{\text{strong}}$, $y$ is the ground-truth label, and $\lambda$ controls the weight of the consistency term. The two classification losses ensure that the model maintains discriminative capability under both weak and strong corruption. The consistency loss further encourages the model to produce aligned predictions across the two severity levels, reinforcing robustness against corruption intensity variation.

\begin{table*}[t]
\centering
\caption{Configuration of the bitstream corruption dataset and the resulting decode rates. Each scenario is defined by a fixed parameter tuple $(S, P, p_f, q)$. The effective corruption ratio $\rho \approx P \cdot q$ indicates the expected proportion of corrupted bytes. Decode Rate (\%) denotes the valid decoding rates of standard image decoders on the Stanford40, PPMI, and PASCAL VOC 2012 datasets, respectively.}
\label{tab:benchmark}
\renewcommand{\arraystretch}{1.15}
\setlength{\tabcolsep}{6pt} 
\begin{tabular}{l c c c c c c c c}
\toprule
\multirow{2}{*}{\textbf{Scenario}} & \multirow{2}{*}{$S$} & \multirow{2}{*}{$P$} & \multirow{2}{*}{$p_f$} & \multirow{2}{*}{$q$} & \multirow{2}{*}{$\rho$ (\%)} & \multicolumn{3}{c}{\textbf{Decode Rate (\%)}} \\
\cmidrule(lr){7-9}
 & & & & & & Stanford40 & PPMI & VOC2012 \\
\midrule
intact        & --  & --   & --  & --   & 0     & 100.0 & 100.0 & 100.0 \\
\midrule
Light-Flip   & 128 & 0.35 & 1.0 & 0.35 & 12.25 & 0.45  & 0.62  & 0.38  \\
Light-Loss   & 128 & 0.35 & 0.0 & 0.35 & 12.25 & 1.07  & 1.05  & 1.13  \\
Light-Mixed  & 128 & 0.35 & 0.5 & 0.35 & 12.25 & 0.56  & 0.62  & 0.75  \\
\midrule
Medium-Flip  & 64  & 0.55 & 1.0 & 0.55 & 30.25 & 0.04  & 0.05  & 0.15  \\
Medium-Loss  & 64  & 0.55 & 0.0 & 0.55 & 30.25 & 0.05  & 0.03  & 0.01  \\
Medium-Mixed & 64  & 0.55 & 0.5 & 0.55 & 30.25 & 0.05  & 0.10  & 0.02  \\
\midrule
Heavy-Flip   & 32  & 0.65 & 1.0 & 0.65 & 42.25 & 0.00  & 0.00  & 0.00  \\
Heavy-Loss   & 32  & 0.65 & 0.0 & 0.65 & 42.25 & 0.00  & 0.00  & 0.00  \\
Heavy-Mixed  & 32  & 0.65 & 0.5 & 0.65 & 42.25 & 0.00  & 0.00  & 0.00  \\
\midrule
Extreme-Flip & 16  & 0.75 & 1.0 & 0.75 & 56.25 & 0.00  & 0.00  & 0.00  \\
Extreme-Loss & 16  & 0.75 & 0.0 & 0.75 & 56.25 & 0.00  & 0.00  & 0.00  \\
Extreme-Mixed& 16  & 0.75 & 0.5 & 0.75 & 56.25 & 0.00  & 0.00  & 0.00  \\
\bottomrule
\end{tabular}
\end{table*}

\section{Experiments}\label{sec:experiments}

\subsection{Experimental Setup}\label{subsec:setup}

\subsubsection{Evaluation Datasets and Metrics}

We evaluate all methods on image bitstreams from Stanford40, PPMI, and PASCAL VOC 2012 Action. Following the Real-world Bitstream Corruption Simulator (RBCS) protocol introduced in~\cite{li2026bitstream}, each image is represented as a raw byte sequence and corrupted using a four-parameter process defined by $(S,P,p_f,q)$. Specifically, the byte sequence is first divided into non-overlapping segments of size $S$, and each segment is selected for corruption with probability $P$. For each selected segment, Bit-Flip is applied with probability $p_f$, while Byte-Loss is applied with probability $1-p_f$. The parameter $q$ controls the probability that each byte within the selected segment is corrupted. Bit-Flip changes byte values while preserving the sequence length, whereas Byte-Loss removes bytes and shifts the positions of subsequent bytes.

For Stanford40 and PPMI, we follow the 13-setting BAR-D evaluation protocol, and the same corruption configurations are further applied to PASCAL VOC 2012 Action. The evaluation includes one intact setting and twelve corrupted settings, obtained by combining four corruption severity levels (Light, Medium, Heavy, and Extreme) with three corruption types (Flip, Loss, and Mixed). Table~\ref{tab:benchmark} lists the parameter configurations and decoding success rates for all settings.

We report Top-1 Accuracy (\%) and mean Average Precision (mAP, \%) as evaluation metrics. For methods that depend on image decoding, samples that fail to decode are treated as misclassifications with zero prediction confidence. All corruption tests use a fixed random seed for reproducibility.

\subsubsection{SOTA Methods in Different Domains}

We compare ByteAction against state-of-the-arts (SOTA) methods from pixel, compressed, and bitstream domains.

\noindent\textbf{Pixel-domain methods} require full JPEG decoding prior to recognition. We evaluate five widely adopted architectures: ResNet-50, ViT-B/16, Swin-Tiny, ConvNeXt-Tiny, and DeiT-Small.

\noindent\textbf{Compressed-domain methods} operate on partially decoded DCT coefficients while still requiring structural parsing of the JPEG bitstream. We evaluate Optimal Codebook \cite{Jamil2019CBIR}, Direct Feature Extractor \cite{Shen1996Edge}, and Transform-Bitstream Classifier \cite{hill2021transform}.

\noindent\textbf{Bitstream-domain methods} process raw image bitstream without any decoding step. We evaluate five models: bGPT \cite{Wu2024bGPT}, MegaByte \cite{yu2023megabyte}, MambaByte \cite{wang2024mambabyte}, MBLM \cite{egli2025multiscale}, and ByteFormer \cite{horton2023bytes}. ByteFormer serves as the backbone of our proposed ByteAction.

\subsubsection{Implementation Details}

All methods share a unified preprocessing pipeline. At test time, each image is resized to $256 \times 256$, center-cropped to $224 \times 224$, and re-encoded as JPEG at quality 100 to produce a intact byte stream, which is then corrupted with a fixed random seed. Pixel-domain models decode the corrupted bytes into RGB images, compressed-domain models extract DCT coefficients through partial decoding, and bitstream-domain models consume the raw bytes directly, truncated or zero-padded to $N = 50{,}000$.

All pixel-domain models are initialized with ImageNet-pretrained weights and fine-tuned on each dataset using AdamW with learning rate $1 \times 10^{-4}$ and weight decay 0.05. Compressed-domain models follow the training configurations from their original papers. For bitstream-domain models, bGPT, MegaByte, MambaByte, and MBLM are adapted for classification following their respective codebases. ByteFormer, and ByteAction all use the Tiny configuration ($d = 192$, 12 Transformer blocks, $k = 8$).

Our method is initialized from the ImageNet-pretrained ByteFormer weights and fine-tuned end-to-end on each dataset using AdamW with learning rate $3 \times 10^{-5}$ and weight decay 0.05. During training, each input bitstream is corrupted at two severity levels: the weak branch samples parameters from $P \in [0.01, 0.25]$ and $q \in [0.01, 0.25]$, while the strong branch samples from $P \in [0.25, 0.50]$ and $q \in [0.25, 0.50]$, with segment size $S$ uniformly sampled from $\{16, 32, 64, 128, 256, 512, 1024\}$ and flip ratio $p_f$ from $[0, 1]$. For BPA, we set $r_{\min} = 0.1, r_{\max} = 0.3, s_{\min} = 0.05, s_{\max} = 0.2, p_{\text{stripe}} = 0.5, K \in [1, 3], p_r = 0.5$, and $p_{\text{BPA}} = 0.5$.The consistency loss uses temperature $T = 2$ and weight $\lambda = 0.5$. Batch size is 32 to accommodate the dual-branch forward pass. All experiments are conducted on a single NVIDIA RTX 4090 GPU.

\begin{table}
\centering
\caption{Comprehensive comparison on Stanford40, PPMI, and VOC2012 Action under the corrupted bitstream dataset (Top-1 Accuracy \%). We report performance for intact inputs and the average across three corruption types at each severity level. ``Corrupt Avg'' denotes the overall mean across all 12 corruption scenarios. Best results are in \textbf{bold}.}
\label{tab:main_results_acc}
\renewcommand{\arraystretch}{0.95} 
\setlength{\tabcolsep}{4.5pt}
\small 
\begin{tabular}{c|l|c|cccccc}
\hline
\textbf{Dataset} & \multicolumn{1}{c|}{Method} & Domain & Intact & \makecell{Light \\ Avg} & \makecell{Medium \\ Avg} & \makecell{Heavy \\ Avg} & \makecell{Extreme \\ Avg} & \makecell{Corrupt \\ Avg} \\
\hline
\multirow{14}{*}{\textbf{Stanford40}} 
& ResNet-50                      & Pixel      & 85.30 & 0.02  & 0.00  & 0.00  & 0.00  & 0.01  \\
& ViT-B/16                       & Pixel      & 83.33 & 0.00  & 0.00  & 0.00  & 0.00  & 0.00  \\
& Swin-Tiny                      & Pixel      & 85.52 & 0.01  & 0.00  & 0.00  & 0.00  & 0.00  \\
& ConvNeXt-Tiny                  & Pixel      & \textbf{87.55} & 0.02  & 0.00  & 0.00  & 0.00  & 0.00  \\
& DeiT-Small                     & Pixel      & 81.36 & 0.00  & 0.00  & 0.00  & 0.00  & 0.00  \\
\cline{2-9}
& Optimal Codebook                & Compressed & 12.51 & 0.09  & 0.01  & 0.00  & 0.00  & 0.02  \\
& Direct Feature Extractor        & Compressed & 19.14 & 0.13  & 0.01  & 0.00  & 0.00  & 0.04  \\
& Transform-Bitstream Classifier & Compressed & 29.32 & 0.20  & 0.01  & 0.00  & 0.00  & 0.05  \\
\cline{2-9}
& bGPT                            & Bitstream  & 6.54  & 4.86  & 4.19  & 4.04  & 3.83  & 4.23  \\
& Megabyte                        & Bitstream  & 4.75  & 4.19  & 3.93  & 3.75  & 3.55  & 3.85  \\
& Mambabyte                       & Bitstream  & 3.38  & 3.38  & 3.38  & 3.38  & 3.38  & 3.38  \\
& MBLM                            & Bitstream  & 3.24  & 3.07  & 2.77  & 2.66  & 2.62  & 2.78  \\
& ByteFormer                      & Bitstream  & 61.33 & 53.38 & 25.64 & 12.50 & 5.91  & 24.36 \\
& \textbf{ByteAction (Ours)}       & Bitstream  & 62.00 & \textbf{59.32} & \textbf{50.62} & \textbf{41.70} & \textbf{26.75} & \textbf{44.60} \\
\hline
\multirow{14}{*}{\textbf{PPMI}} 
& ResNet-50                      & Pixel      & 71.65 & 0.07  & 0.00  & 0.00  & 0.00  & 0.02  \\
& ViT-B/16                       & Pixel      & 64.84 & 0.00  & 0.00  & 0.00  & 0.00  & 0.00  \\
& Swin-Tiny                      & Pixel      & 67.03 & 0.00  & 0.00  & 0.00  & 0.00  & 0.00  \\
& ConvNeXt-Tiny                  & Pixel      & \textbf{74.56} & 0.07  & 0.00  & 0.00  & 0.00  & 0.02  \\
& DeiT-Small                     & Pixel      & 65.75 & 0.05  & 0.00  & 0.00  & 0.00  & 0.01  \\
\cline{2-9}
& Optimal Codebook                & Compressed & 9.29  & 0.07  & 0.00  & 0.00  & 0.00  & 0.02  \\
& Direct Feature Extractor        & Compressed & 7.81  & 0.06  & 0.00  & 0.00  & 0.00  & 0.02  \\
& Transform-Bitstream Classifier & Compressed & 17.53 & 0.13  & 0.01  & 0.00  & 0.00  & 0.04  \\
\cline{2-9}
& bGPT                            & Bitstream  & 12.15 & 9.86  & 9.00  & 8.00  & 7.43  & 8.58  \\
& Megabyte                        & Bitstream  & 12.29 & 10.56 & 9.27  & 8.24  & 7.32  & 8.85  \\
& Mambabyte                       & Bitstream  & 6.38  & 6.32  & 5.75  & 5.43  & 4.70  & 5.55  \\
& MBLM                            & Bitstream  & 12.91 & 12.34 & 12.10 & 11.94 & 10.45 & 11.71 \\
& ByteFormer                      & Bitstream  & 50.07 & 37.19 & 15.28 & 8.23  & 4.99  & 16.42 \\
& \textbf{ByteAction (Ours)}       & Bitstream  & 49.83 & \textbf{44.07} & \textbf{35.03} & \textbf{27.90} & \textbf{18.07} & \textbf{31.27} \\
\hline
\multirow{14}{*}{\textbf{VOC2012}} 
& ResNet-50                      & Pixel      & 71.29 & 0.05  & 0.00  & 0.00  & 0.00  & 0.01  \\
& ViT-B/16                       & Pixel      & 73.97 & 0.08  & 0.00  & 0.00  & 0.00  & 0.02  \\
& Swin-Tiny                      & Pixel      & 74.87 & 0.06  & 0.00  & 0.00  & 0.00  & 0.02  \\
& ConvNeXt-Tiny                  & Pixel      & \textbf{77.46} & 0.06  & 0.00  & 0.00  & 0.00  & 0.02  \\
& DeiT-Small                     & Pixel      & 75.06 & 0.05  & 0.00  & 0.00  & 0.00  & 0.01  \\
\cline{2-9}
& Optimal Codebook                & Compressed & 37.89 & 0.04  & 0.00  & 0.00  & 0.00  & 0.01  \\
& Direct Feature Extractor        & Compressed & 36.91 & 0.06  & 0.00  & 0.00  & 0.00  & 0.02  \\
& Transform-Bitstream Classifier & Compressed & 38.00 & 0.03  & 0.00  & 0.00  & 0.00  & 0.01  \\
\cline{2-9}
& bGPT                            & Bitstream  & 17.49 & 15.30 & 13.73 & 13.41 & 12.50 & 13.74 \\
& Megabyte                        & Bitstream  & 18.02 & 18.41 & 18.70 & 18.83 & 18.07 & 18.50 \\
& Mambabyte                       & Bitstream  & 11.36 & 11.76 & 12.03 & 12.31 & 12.74 & 12.21 \\
& MBLM                            & Bitstream  & 18.70 & 18.54 & 18.75 & 18.57 & 18.24 & 18.52 \\
& ByteFormer                      & Bitstream  & 59.07 & 52.15 & 27.78 & 19.93 & 14.88 & 28.68 \\
& \textbf{ByteAction (Ours)}       & Bitstream  & 60.46 & \textbf{58.93} & \textbf{52.80} & \textbf{44.43} & \textbf{32.66} & \textbf{47.20} \\
\hline
\end{tabular}
\end{table}

\begin{table}
\centering
\caption{Comprehensive comparison on Stanford40, PPMI, and VOC2012 Action under corrupted bitstream dataset (mAP \%). We report performance for intact inputs and the average across three corruption types at each severity level. ``Corrupt Avg'' denotes the overall mean across all 12 corruption scenarios. Best results are in \textbf{bold}.}
\label{tab:main_results_map}
\renewcommand{\arraystretch}{0.95} 
\setlength{\tabcolsep}{4.5pt}
\small 
\begin{tabular}{c|l|c|cccccc}
\hline
\textbf{Dataset} & \multicolumn{1}{c|}{Method} & Domain & Intact & \makecell{Light \\ Avg} & \makecell{Medium \\ Avg} & \makecell{Heavy \\ Avg} & \makecell{Extreme \\ Avg} & \makecell{Corrupt \\ Avg} \\
\hline
\multirow{14}{*}{\textbf{Stanford40}} 
& ResNet-50                      & Pixel      & 90.09 & 2.57  & 2.52  & 2.50  & 2.50  & 2.52  \\
& ViT-B/16                       & Pixel      & 86.23 & 2.59  & 2.51  & 2.50  & 2.50  & 2.52  \\
& Swin-Tiny                      & Pixel      & 90.08 & 2.58  & 2.52  & 2.50  & 2.50  & 2.52  \\
& ConvNeXt-Tiny                  & Pixel      & \textbf{92.66} & 2.58  & 2.52  & 2.50  & 2.50  & 2.52  \\
& DeiT-Small                     & Pixel      & 85.94 & 2.57  & 2.52  & 2.50  & 2.50  & 2.52  \\
\cline{2-9}
& Optimal Codebook                & Compressed & 10.30 & 2.74  & 2.72  & 2.50  & 2.50  & 2.61  \\
& Direct Feature Extractor        & Compressed & 15.28 & 2.55  & 2.51  & 2.50  & 2.50  & 2.52  \\
& Transform-Bitstream Classifier & Compressed & 28.34 & 2.85  & 2.62  & 2.50  & 2.50  & 2.62  \\
\cline{2-9}
& bGPT                            & Bitstream  & 4.52  & 3.97  & 3.80  & 3.67  & 3.54  & 3.74  \\
& Megabyte                        & Bitstream  & 4.08  & 3.84  & 3.82  & 3.85  & 3.74  & 3.81  \\
& Mambabyte                       & Bitstream  & 3.03  & 3.00  & 3.07  & 3.05  & 3.03  & 3.04  \\
& MBLM                            & Bitstream  & 3.54  & 3.51  & 3.51  & 3.50  & 3.50  & 3.51  \\
& ByteFormer                      & Bitstream  & 63.84 & 55.12 & 25.34 & 12.16 & 5.53  & 24.54 \\
& \textbf{ByteAction (Ours)}       & Bitstream  & 64.49 & \textbf{61.13} & \textbf{51.06} & \textbf{40.28} & \textbf{24.04} & \textbf{44.13} \\
\hline
\multirow{14}{*}{\textbf{PPMI}} 
& ResNet-50                      & Pixel      & 77.21 & 4.30  & 4.22  & 4.17  & 4.17  & 4.22  \\
& ViT-B/16                       & Pixel      & 66.63 & 4.31  & 4.21  & 4.17  & 4.17  & 4.22  \\
& Swin-Tiny                      & Pixel      & 73.02 & 4.35  & 4.21  & 4.17  & 4.17  & 4.22  \\
& ConvNeXt-Tiny                  & Pixel      & \textbf{81.24} & 4.35  & 4.21  & 4.17  & 4.17  & 4.23  \\
& DeiT-Small                     & Pixel      & 70.95 & 4.34  & 4.21  & 4.17  & 4.17  & 4.22  \\
\cline{2-9}
& Optimal Codebook                & Compressed & 7.61  & 4.28  & 4.19  & 4.17  & 4.17  & 4.20  \\
& Direct Feature Extractor        & Compressed & 6.95  & 4.25  & 4.19  & 4.17  & 4.17  & 4.20  \\
& Transform-Bitstream Classifier & Compressed & 16.69 & 4.45  & 4.22  & 4.17  & 4.17  & 4.25  \\
\cline{2-9}
& bGPT                            & Bitstream  & 11.15 & 10.06 & 9.50  & 9.05  & 8.68  & 9.32  \\
& Megabyte                        & Bitstream  & 11.51 & 10.61 & 9.50  & 9.07  & 8.42  & 9.40  \\
& Mambabyte                       & Bitstream  & 6.18  & 6.24  & 6.31  & 6.39  & 6.37  & 6.33  \\
& MBLM                            & Bitstream  & 11.25 & 11.20 & 11.14 & 11.10 & 10.52 & 10.99 \\
& ByteFormer                      & Bitstream  & 52.37 & 40.20 & 16.42 & 9.03  & 6.14  & 17.95 \\
& \textbf{ByteAction (Ours)}       & Bitstream  & 51.29 & \textbf{44.98} & \textbf{34.97} & \textbf{25.93} & \textbf{16.05} & \textbf{30.48} \\
\hline
\multirow{14}{*}{\textbf{VOC2012}} 
& ResNet-50                      & Pixel      & 80.52 & 10.11 & 10.03 & 10.00 & 10.00 & 10.04 \\
& ViT-B/16                       & Pixel      & 82.44 & 10.12 & 10.03 & 10.00 & 10.00 & 10.04 \\
& Swin-Tiny                      & Pixel      & 84.42 & 10.07 & 10.04 & 10.00 & 10.00 & 10.03 \\
& ConvNeXt-Tiny                  & Pixel      & \textbf{85.85} & 10.11 & 10.04 & 10.00 & 10.00 & 10.04 \\
& DeiT-Small                     & Pixel      & 83.76 & 10.09 & 10.03 & 10.00 & 10.00 & 10.03 \\
\cline{2-9}
& Optimal Codebook                & Compressed & 36.48 & 10.09 & 10.02 & 10.00 & 10.00 & 10.03 \\
& Direct Feature Extractor        & Compressed & 34.74 & 10.04 & 10.01 & 10.00 & 10.00 & 10.01 \\
& Transform-Bitstream Classifier & Compressed & 39.02 & 10.02 & 10.01 & 10.00 & 10.00 & 10.01 \\
\cline{2-9}
& bGPT                            & Bitstream  & 15.33 & 15.08 & 14.62 & 14.06 & 13.93 & 14.42 \\
& Megabyte                        & Bitstream  & 16.74 & 16.91 & 16.96 & 16.94 & 16.92 & 16.93 \\
& Mambabyte                       & Bitstream  & 16.49 & 16.60 & 16.52 & 16.56 & 16.53 & 16.55 \\
& MBLM                            & Bitstream  & 17.66 & 17.63 & 17.55 & 17.40 & 17.35 & 17.48 \\
& ByteFormer                      & Bitstream  & 67.34 & 59.83 & 38.06 & 27.93 & 20.68 & 36.63 \\
& \textbf{ByteAction (Ours)}       & Bitstream  & 69.13 & \textbf{66.92} & \textbf{59.07} & \textbf{49.83} & \textbf{35.13} & \textbf{52.74} \\
\hline
\end{tabular}
\end{table}

\begin{figure}
  \centering
  \begin{subfigure}[b]{\linewidth}
    \centering
    \begin{overpic}[width=0.92\linewidth,trim=0 0 0 0,clip]{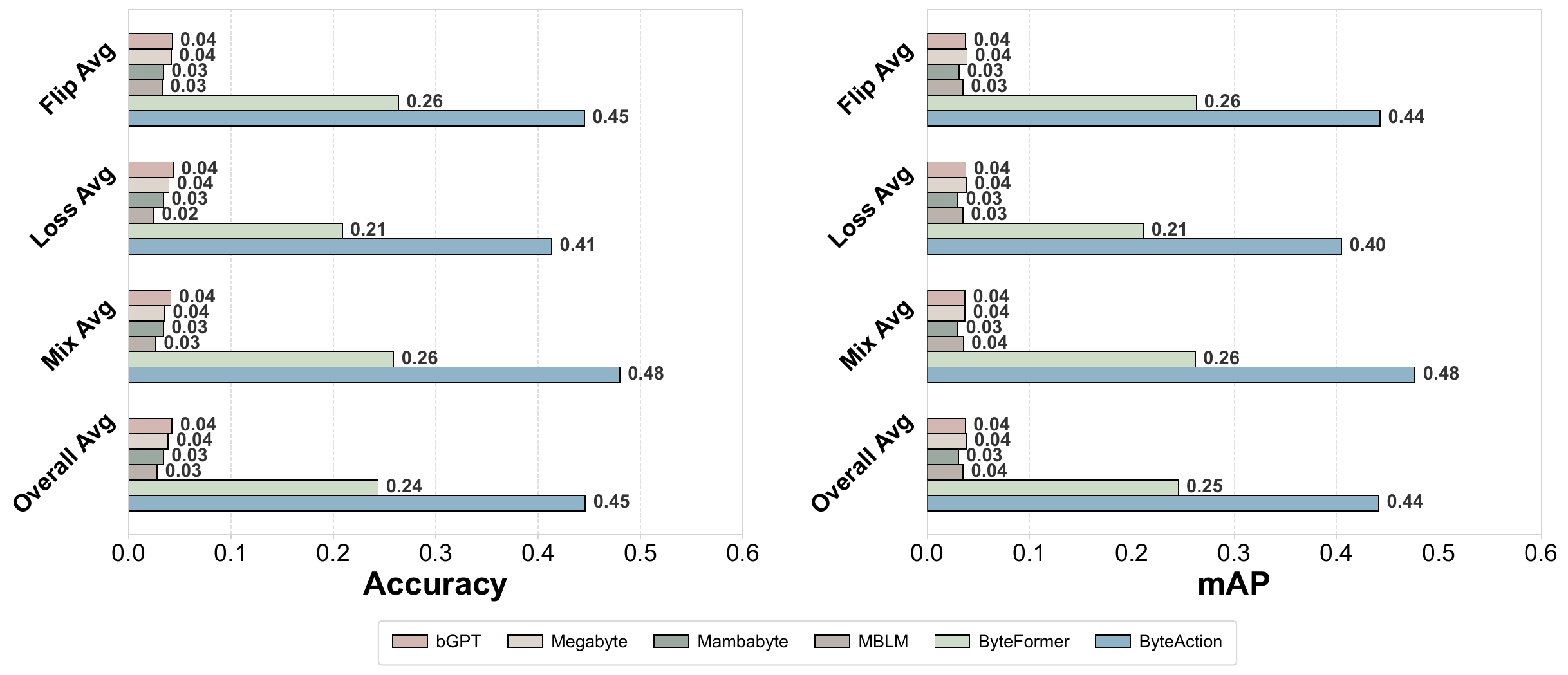}
      \put(-4,40){\bfseries\small (a)}
    \end{overpic}
  \end{subfigure}

  \begin{subfigure}[b]{\linewidth}
    \centering
    \begin{overpic}[width=0.92\linewidth,trim=0 0 0 0,clip]{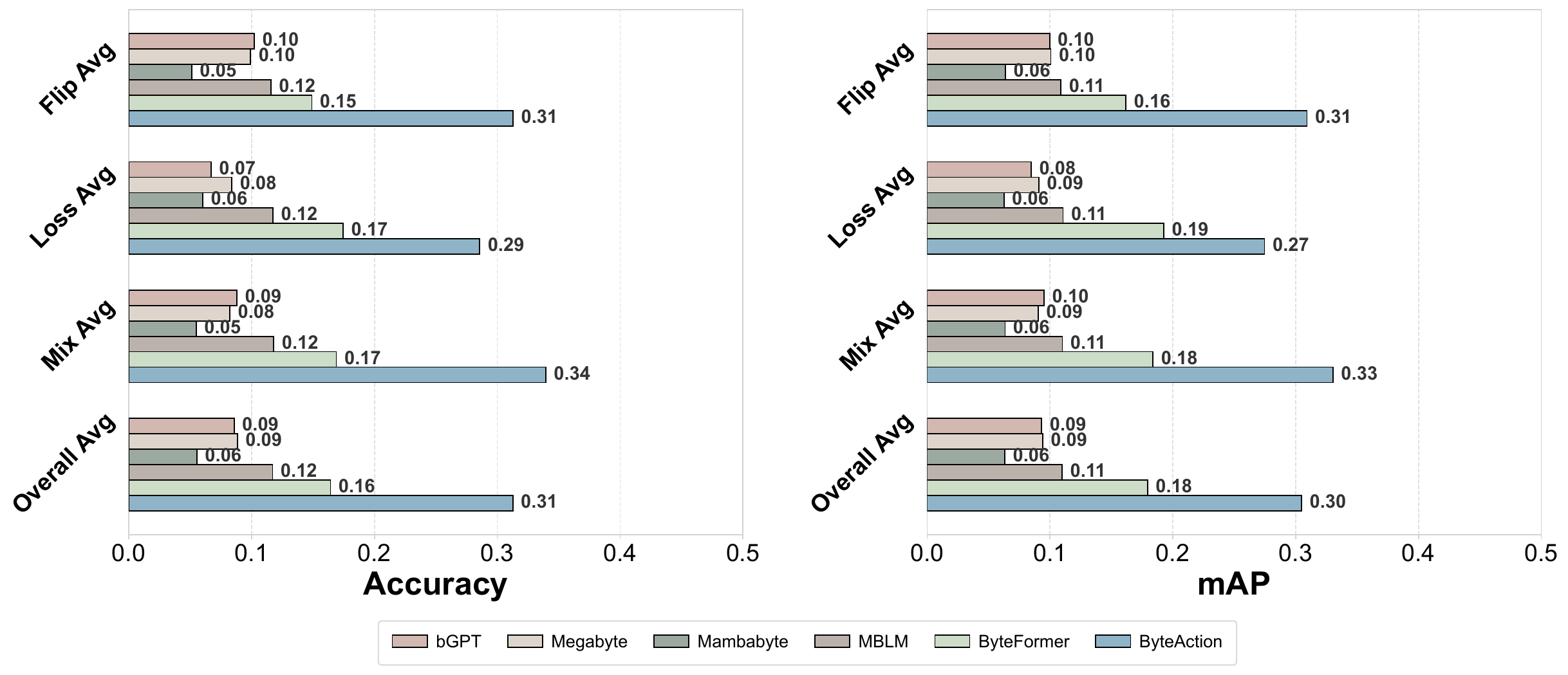}
      \put(-4,40){\bfseries\small (b)}
    \end{overpic}
  \end{subfigure}

  \begin{subfigure}[b]{\linewidth}
    \centering
    \begin{overpic}[width=0.92\linewidth,trim=0 0 0 0,clip]{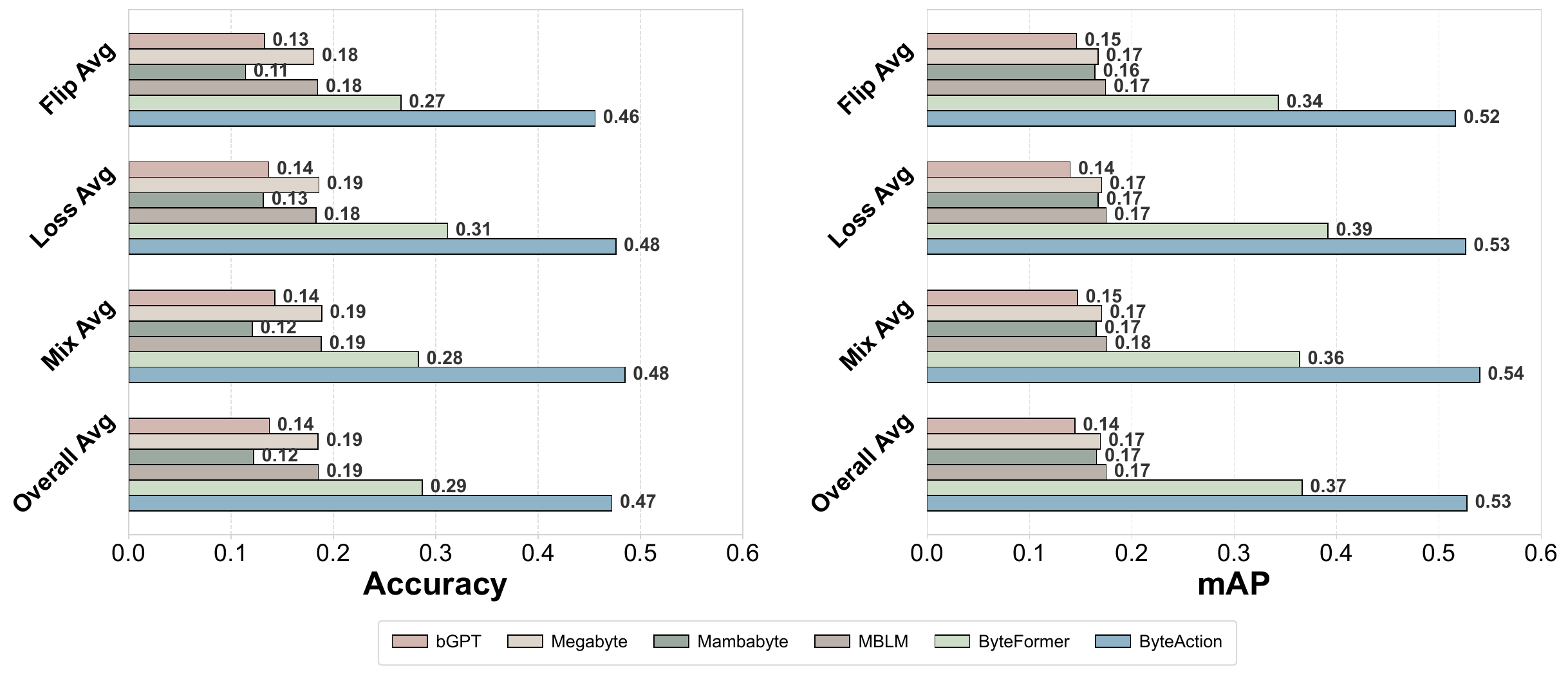}
      \put(-4,40){\bfseries\small (c)}
    \end{overpic}
  \end{subfigure}
  \caption{Comparison of bitstream-domain methods by corruption type on (a) Stanford40, (b) PPMI, and (c) VOC2012 Action. Each group reports the average Accuracy (left) and mAP (right) across four severity levels within the corresponding corruption type (Flip, Loss, Mixed) and overall. }
  \label{fig:Different_Corrupt_type}
\end{figure}

\subsection{Comparison with SOTA Methods in Different Domains}\label{subsec:comparison}

\subsubsection{Overall Results}

Table~\ref{tab:main_results_acc} and Table~\ref{tab:main_results_map} present the Top-1 Accuracy and mAP of all compared methods on the corrupted bitstream dataset. Pixel-domain and compressed-domain methods achieve strong intact performance but degrade to random-chance levels once corruption is introduced, as the JPEG decoding process fails under damaged bitstreams. We therefore focus our discussion on bitstream-domain methods.

Among bitstream-domain models, general-purpose byte-level models (bGPT, MegaByte, MambaByte, MBLM) achieve limited accuracy even on intact inputs, indicating that their architectures are not well suited for visual recognition from JPEG byte sequences. ByteFormer, designed specifically for byte-level visual understanding, serves as a substantially stronger baseline. However, its performance drops sharply as corruption severity increases, falling from 61.33\% to 5.91\% Accuracy on Stanford40 across the four corruption levels.

ByteAction achieves the highest Corrupt Average on all three datasets in both metrics. On Stanford40, it improves the Corrupt Average Accuracy from 24.36\% (ByteFormer) to 44.60\%, a gain of over 20 percentage points. Similar improvements are observed on PPMI (16.42\% $\to$ 31.27\%) and VOC2012 (28.68\% $\to$ 47.20\%). The advantage is consistent across all severity levels, with the largest gains at Medium and Heavy corruption where sufficient residual information remains for the proposed BPA and consistency training to exploit. Meanwhile, ByteAction maintains competitive intact accuracy on all three datasets, confirming that the corruption robustness is not gained at the expense of raw bitstream understanding capability.

\subsubsection{Analysis by Corruption Type}

Fig.~\ref{fig:Different_Corrupt_type} further decomposes the Corrupt Average by corruption type (Flip, Loss, Mixed) for all bitstream-domain methods on three datasets. ByteAction achieves the highest Accuracy and mAP under all three corruption types across all datasets, confirming that the proposed method is effective regardless of the corruption mechanism.
A consistent pattern is observed across all models: Mixed-type corruption yields higher accuracy than pure Flip or pure Loss at the same overall severity. This is because the Mixed setting distributes its corruption between two mechanisms, resulting in lower per-mechanism intensity and preserving more diverse residual information for the model to exploit. Among the two pure types, Loss is generally more destructive than Flip, as byte deletion not only removes information but also shifts the positions of all subsequent bytes, disrupting the sequential structure that the model relies on. This difference is particularly evident on Stanford40 and VOC2012, where the gap between Flip and Loss is substantial for ByteFormer. ByteAction narrows this gap considerably, indicating that BPA and consistency training together provide robustness against both value corruption and positional displacement.

\begin{figure}
    \centering
    \begin{subfigure}[b]{0.48\textwidth}
        \centering
        \includegraphics[width=\textwidth]{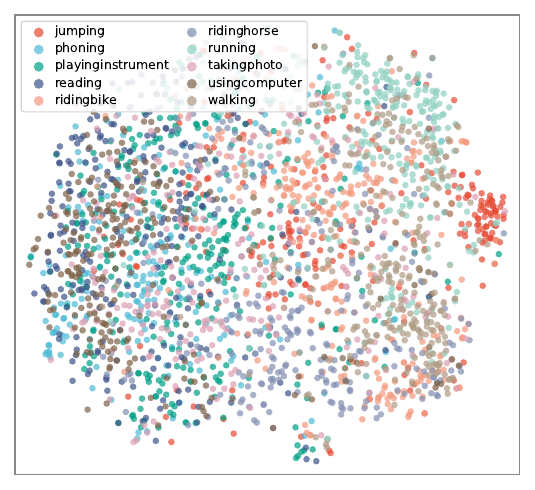}
        \caption{} 
        \label{fig:tsne_bf}
    \end{subfigure}
    \hfill
    \begin{subfigure}[b]{0.48\textwidth}
        \centering
        \includegraphics[width=\textwidth]{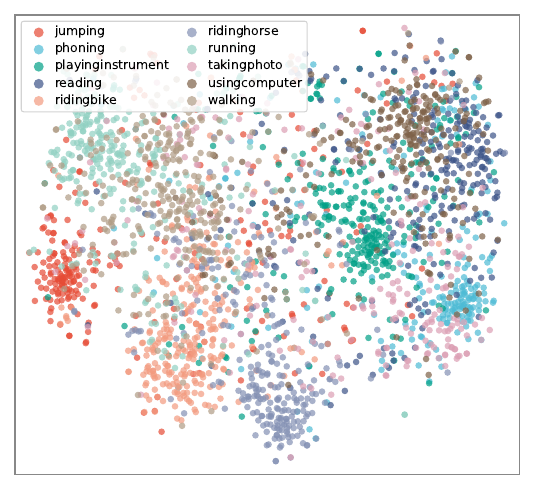}
        \caption{} 
        \label{fig:tsne_bs}
    \end{subfigure}   
    
    \caption{t-SNE visualization of learned features on VOC2012 Action 
    (10 classes) under corruption. (a) ByteFormer. (b) ByteAction (Ours). 
    Best viewed in color.}
    \label{fig:tsne}
\end{figure}

\begin{figure}
    \centering
    \resizebox{0.95\textwidth}{!}{%
    \begin{minipage}{\textwidth}
        \centering

        \begin{subfigure}[b]{0.495\textwidth}
            \centering
            \includegraphics[width=\linewidth]{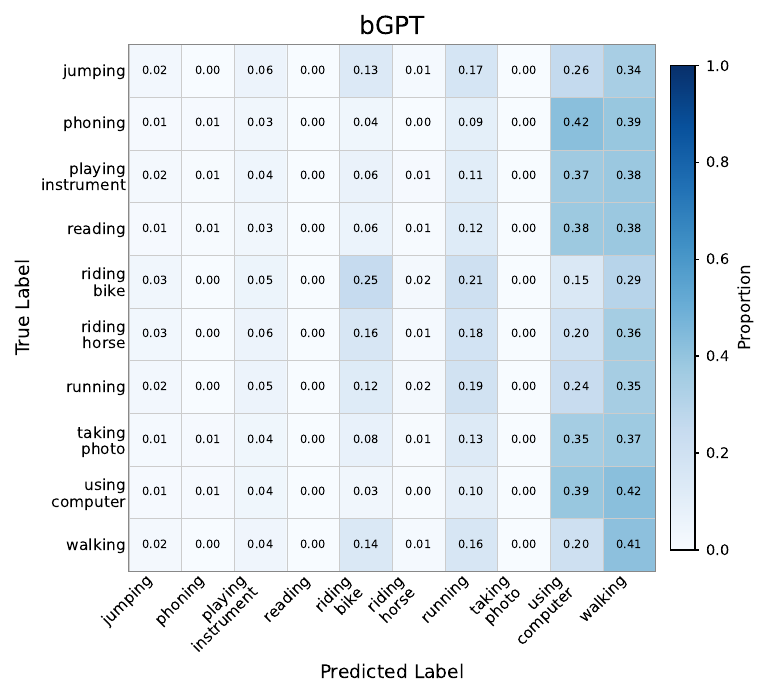}
        \end{subfigure}\hfill
        \begin{subfigure}[b]{0.495\textwidth}
            \centering
            \includegraphics[width=\linewidth]{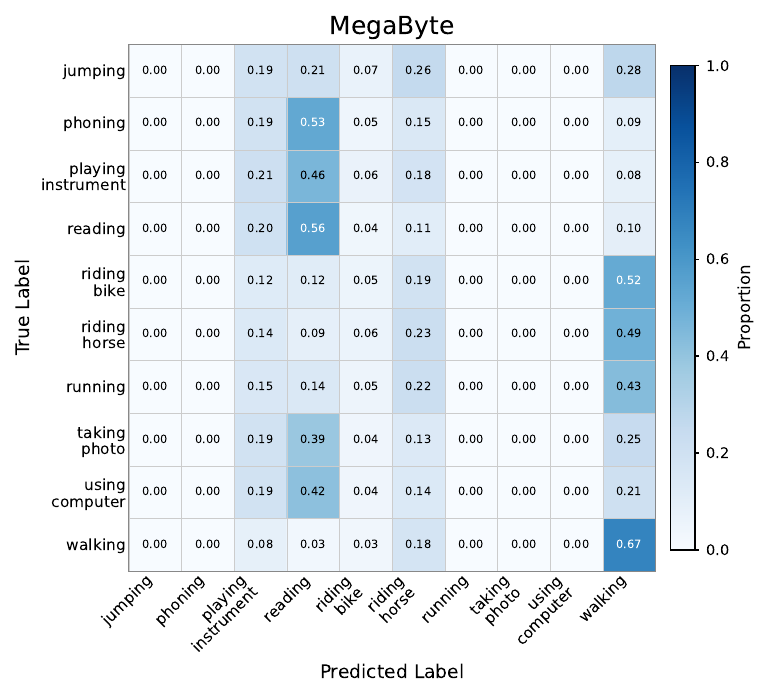}
        \end{subfigure}

        \vspace{-2mm}

        \begin{subfigure}[b]{0.495\textwidth}
            \centering
            \includegraphics[width=\linewidth]{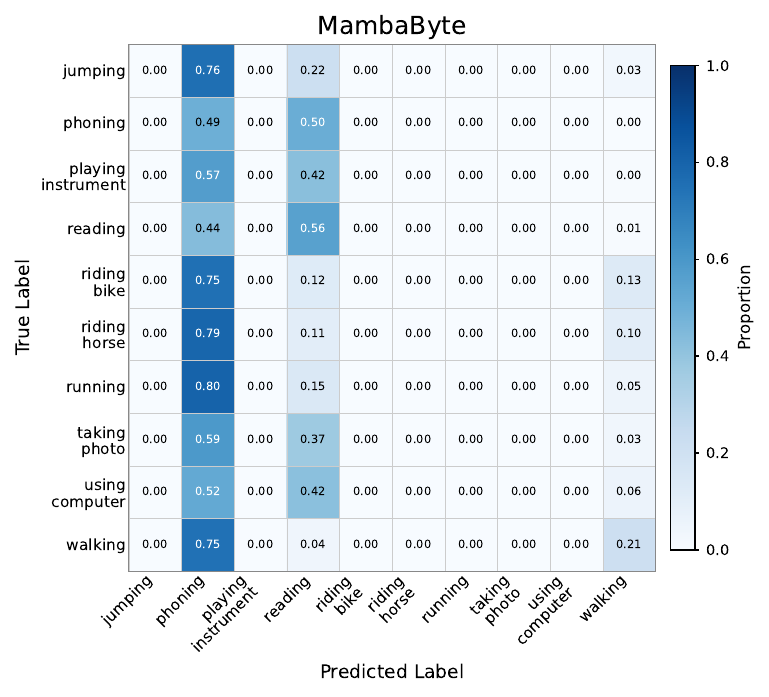}
        \end{subfigure}\hfill
        \begin{subfigure}[b]{0.495\textwidth}
            \centering
            \includegraphics[width=\linewidth]{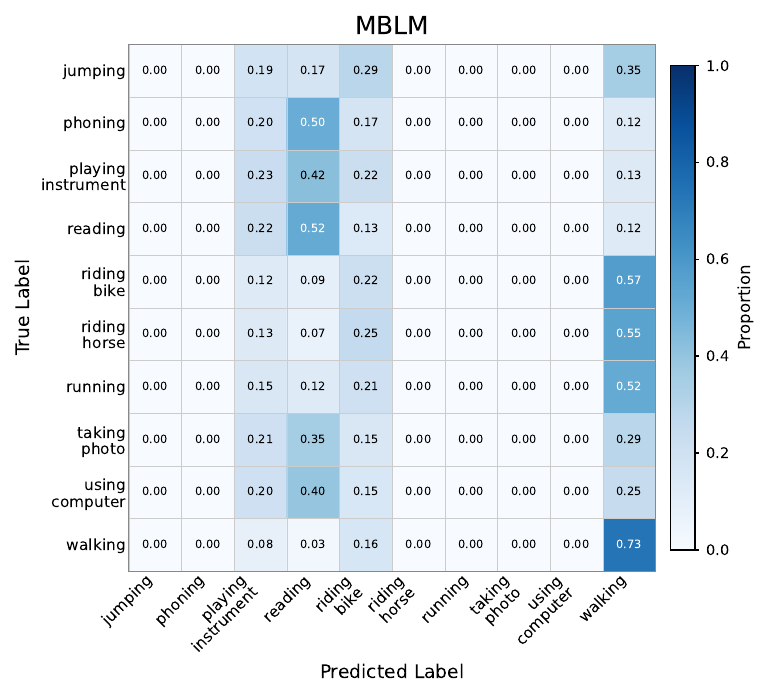}
        \end{subfigure}

        \vspace{-2mm}

        \begin{subfigure}[b]{0.495\textwidth}
            \centering
            \includegraphics[width=\linewidth]{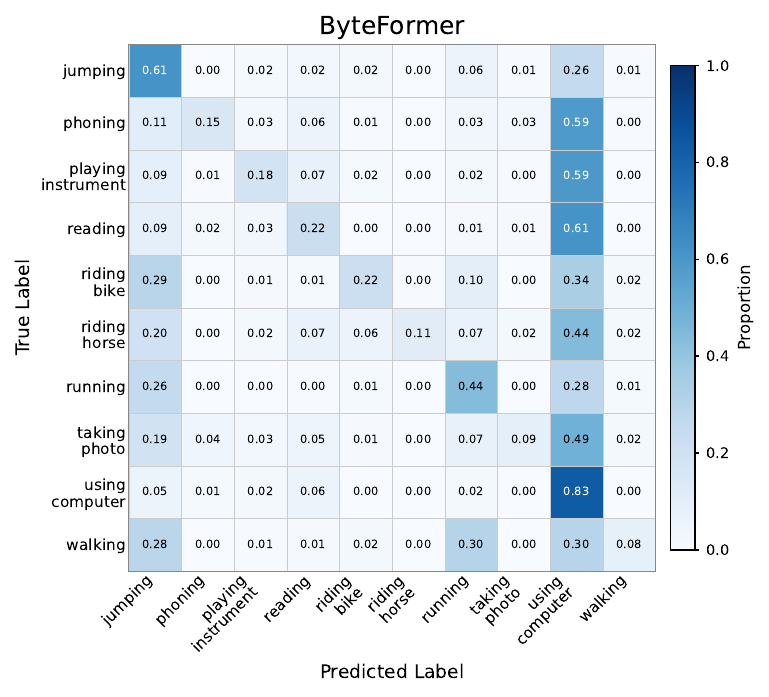}
        \end{subfigure}\hfill
        \begin{subfigure}[b]{0.495\textwidth}
            \centering
            \includegraphics[width=\linewidth]{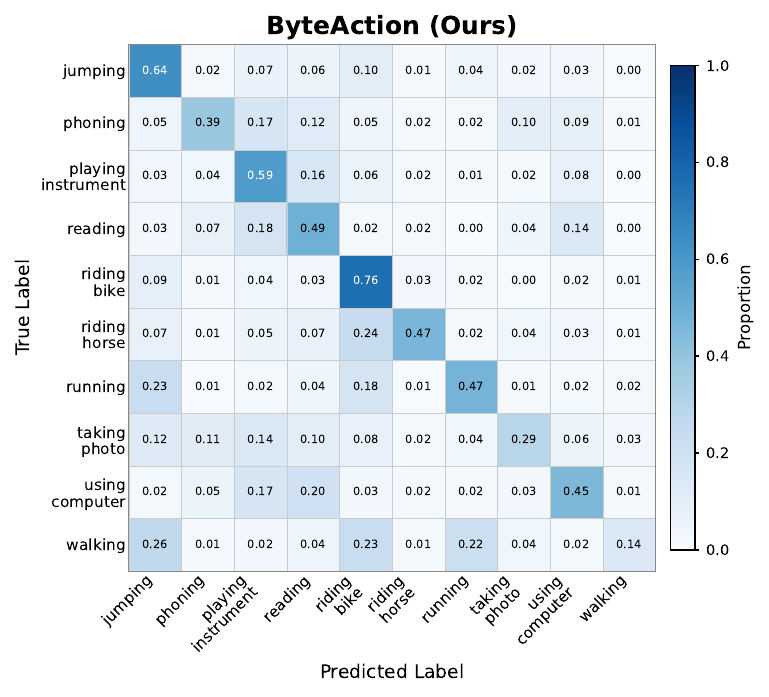}
        \end{subfigure}
    \end{minipage}
    }

    \vspace{-2mm}
    \caption{Confusion matrices of bitstream-domain methods on VOC2012 Action, averaged over all 12 corruption scenarios.}
    \label{fig:confusion}
\end{figure}

\begin{figure}
  \centering

  \begin{subfigure}[b]{\linewidth}
    \centering
    \begin{overpic}[width=1\linewidth,trim=0 0 0 0,clip]{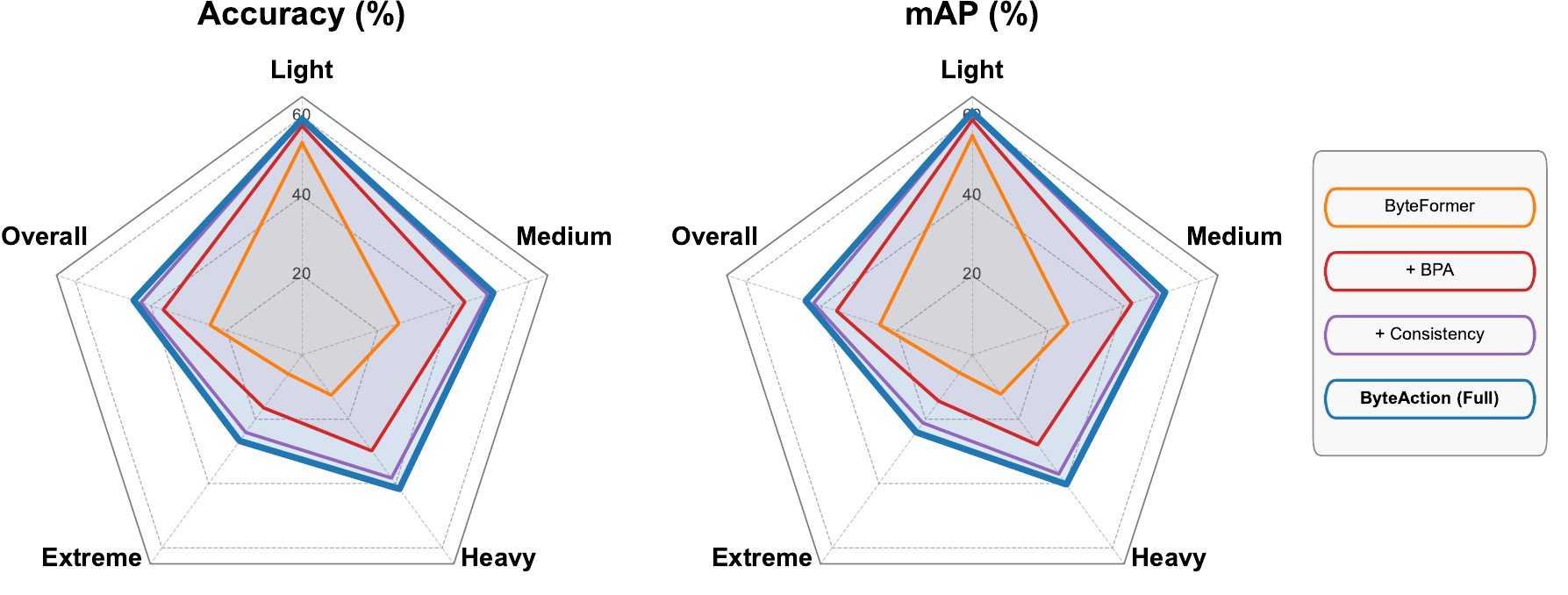}
      \put(0,36){\bfseries\small (a)}
    \end{overpic}
  \end{subfigure}

  \vspace{0.5em}

  \begin{subfigure}[b]{\linewidth}
    \centering
    \begin{overpic}[width=1\linewidth,trim=0 0 0 0,clip]{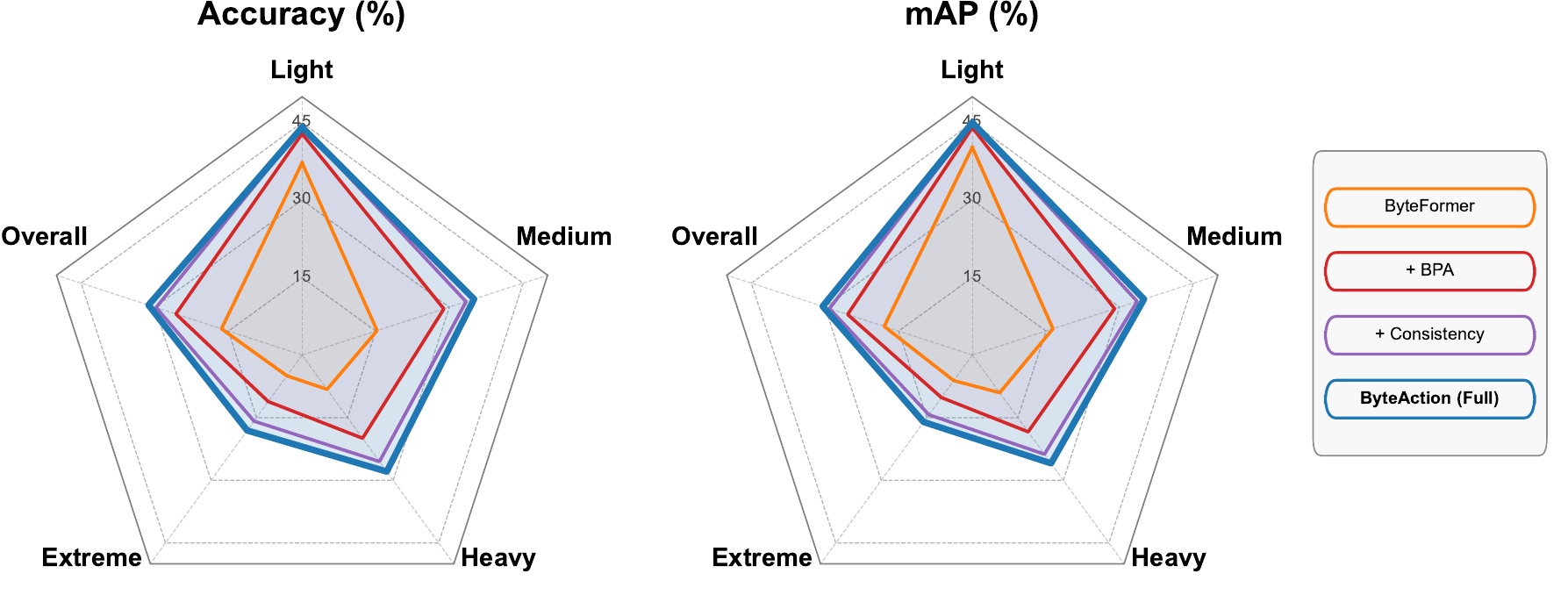}
      \put(0,36){\bfseries\small (b)}
    \end{overpic}
  \end{subfigure}

  \vspace{0.5em}

  \begin{subfigure}[b]{\linewidth}
    \centering
    \begin{overpic}[width=1\linewidth,trim=0 0 0 0,clip]{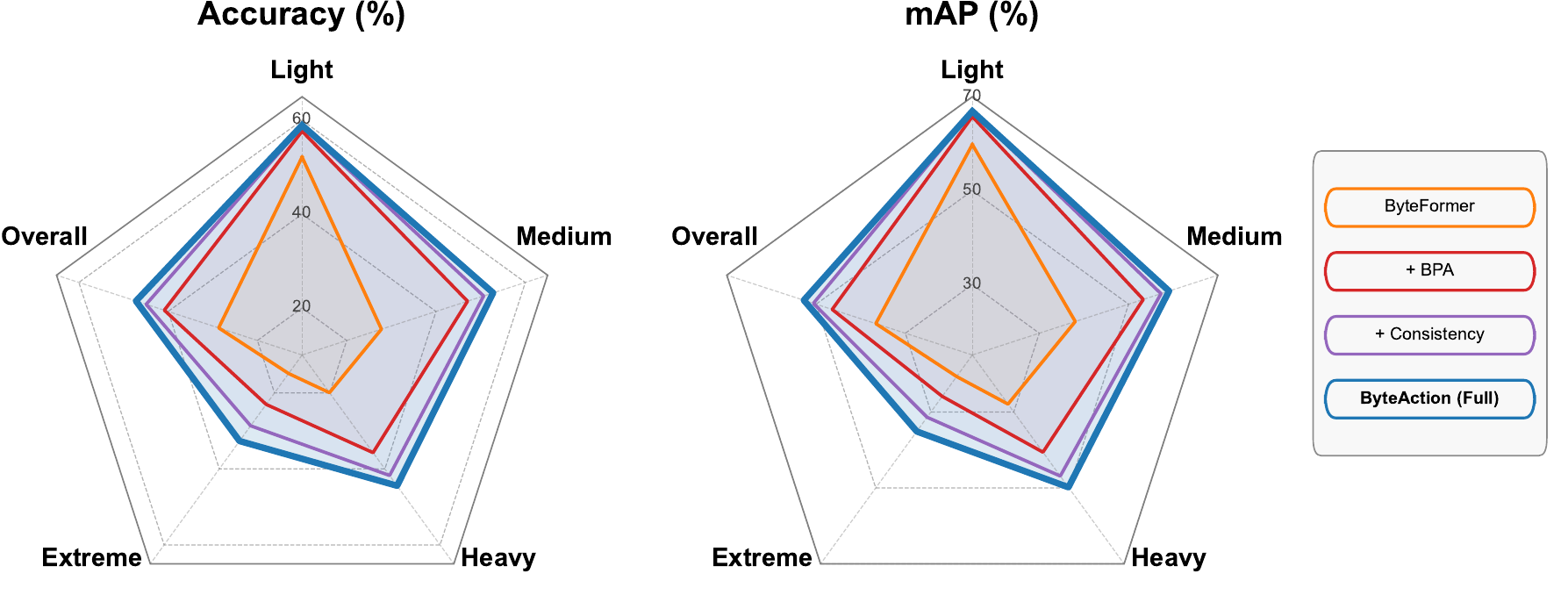}
      \put(0,36){\bfseries\small (c)}
    \end{overpic}
  \end{subfigure}

  \caption{Ablation study on the proposed components across (a) Stanford40, 
(b) PPMI, and (c) VOC2012 Action. Each axis represents the average 
performance at a given corruption severity level (Light, Medium, Heavy, 
Extreme) and the overall Corrupt Average. Left: Top-1 Accuracy (\%). 
Right: mAP (\%). Larger coverage area indicates stronger robustness. }
  \label{fig:Ablation}
\end{figure}

\subsubsection{Feature and Classification Visualization}

To provide a more intuitive understanding of how ByteAction improves corruption robustness, we visualize the learned feature representations and classification behavior on VOC2012 Action. This dataset is selected for visualization due to its moderate number of classes (10), which allows clear visual interpretation. All visualizations are conducted under Medium-Mixed corruption.

Fig.~\ref{fig:tsne} visualizes the t-SNE projections of global features extracted from the last Transformer layer on VOC2012 Action under corruption. ByteFormer (Fig.~\ref{fig:tsne}(a)) shows heavily interleaved clusters with blurred class boundaries, indicating that the features lose their discriminative structure when the bitstream is damaged. In contrast, ByteAction (Fig.~\ref{fig:tsne}(b)) maintains clearly separated clusters, demonstrating that the proposed training strategy effectively preserves feature discriminability under corruption.

Fig.~\ref{fig:confusion} presents the confusion matrices of all bitstream-domain methods, averaged over the 12 corruption scenarios. General-purpose byte models show weak or absent diagonal patterns, with predictions heavily biased toward a few dominant classes regardless of the true label,indicating 
that these models fail to learn meaningful class distinctions from image bitstream. ByteFormer exhibits a visible diagonal structure, but with notable off-diagonal confusion among visually related actions such as ``riding bike'' and ``riding horse''. ByteAction(Ours) shows the strongest diagonal with substantially reduced inter-class confusion across all categories, confirming that the proposed method produces more discriminative and robust representations under corruption.

\subsection{Ablation Studies}\label{subsec:ablation}

To evaluate the contribution of each component in ByteAction, we conduct ablation studies based on the ByteFormer backbone. As shown in Fig.~\ref{fig:Ablation}, we compare five configurations. 
\emph{ByteFormer} denotes the baseline model trained without bitstream corruption augmentation. 
\emph{+ BPA} adds Bitstream Pattern Augmentation to the ByteFormer baseline, while \emph{+ Consistency} adds Corruption Consistency Training. 
\emph{ByteAction (Full)} combines both BPA and Corruption Consistency Training.

Based on the baseline, adding either BPA or Corruption Consistency Training leads to consistent improvements, showing the effectiveness of both components. When the two components are combined, ByteAction achieves the best overall performance on all three datasets. The improvement over the single-component variants suggests that BPA and Corruption Consistency Training benefit the model from different aspects: BPA enriches corrupted byte patterns during training, while Corruption Consistency Training stabilizes predictions across corruption severities.

\section{Conclusion}\label{sec:conclusion}

This paper presented ByteAction, a foundation model for Byte-space Action Recognition (BAR), which recognizes actions directly from corrupted image bitstreams without pixel decoding. To improve robustness in byte space, ByteAction introduces two key components. First, Bitstream Pattern Augmentation (BPA) reshapes one-dimensional byte sequences into two-dimensional byte matrices and applies region-level erasure to diversify bitstream patterns during training. Second, Corruption Consistency Training aligns the predictions of weakly and strongly corrupted views through bidirectional KL divergence, encouraging stable recognition under different corruption severities.Extensive experiments on Stanford40, PPMI, and PASCAL VOC 2012 Action demonstrate that ByteAction achieves state-of-the-art robustness under diverse bitstream corruption settings while maintaining competitive intact-bitstream performance.These results suggest that learning directly from image bitstreams is a promising direction for decoding-free visual recognition. By avoiding explicit pixel reconstruction, ByteAction is suitable for privacy-sensitive scenarios and can remain reliable under bitstream corruption.











\printcredits

\subsection*{Declaration of competing interest}
The authors declare that they have no known competing financial interests or personal relationships that could have appeared to influence the work reported in this paper.

\section*{Acknowledgments}
This work was supported by the the National Natural Science Foundation of China under Grants 62501246.

\subsection*{Data availability} 
Data will be made available on request.

\bibliographystyle{cas-model2-names}

\bibliography{new-refs}



\end{document}